\documentclass[twoside,11pt]{article}

\usepackage[accepted]{melba}
\usepackage{cite}
\usepackage{graphicx}
\usepackage{textcomp}
\usepackage{amsmath,amssymb,amsfonts}
\usepackage{algorithm,algorithmicx}
\usepackage[noend]{algpseudocode}
\usepackage{graphicx}
\usepackage{textcomp}
\usepackage{multirow}
\usepackage{bm}
\usepackage{url}
\usepackage{hyperref}
\usepackage{color}
\usepackage{rotating}

\DeclareMathOperator*{\argmin}{arg\,min}
\newcommand{\norm}[1]{\left\lVert#1\right\rVert}
\usepackage{booktabs}
\usepackage{xr}
\usepackage{caption}
\usepackage{wrapfig}
\usepackage{sidecap}
\usepackage{subcaption}

\usepackage{amsmath,amsfonts}

\melbaheading{2022:017}{https://www.melba-journal.org/papers/2022:017.html}{2022}{1-25}{03/2022}{06/2022}{Wang, Dalca, and Sabuncu}{}{}

\ShortHeadings{Computing Multiple Image Reconstructions with a Single Hypernetwork}{Wang, Dalca, and Sabuncu}
\firstpageno{1}

\title{Computing Multiple Image Reconstructions \\ with a Single Hypernetwork}

\author{\name Alan Q. Wang \email aw847@cornell.edu \\  
	\addr School of Electrical and Computer Engineering, Cornell Tech, New York, NY, USA \\
	\addr Department of Radiology, Weill Cornell Medical School, New York, NY, USA
	\AND
    \name Adrian V. Dalca \email adalca@mit.edu \\  
	\addr Computer Science and Artificial Intelligence Lab, Massachusetts Institute of Technology, Cambridge, MA, USA \\
	\addr A.A. Martinos Center for Biomedical Imaging, Massachusetts General Hospital, Cambridge, MA, USA
	\AND
	\name Mert R. Sabuncu \email msabuncu@cornell.edu \\
	\addr School of Electrical and Computer Engineering, Cornell Tech, New York, NY, USA \\
	\addr Department of Radiology, Weill Cornell Medical School, New York, NY, USA
}

\begin{document}

\maketitle

\begin{abstract}
Deep learning based techniques achieve state-of-the-art results in a wide range of image reconstruction tasks like compressed sensing.
These methods almost always have hyperparameters, such as the weight coefficients that balance the different terms in the optimized loss function.
The typical approach is to train the model for a hyperparameter setting determined with some empirical or theoretical justification.
Thus, at inference time, the model can only compute reconstructions corresponding to the pre-determined hyperparameter values.
In this work, we present a hypernetwork-based approach, called HyperRecon, to train reconstruction models that are agnostic to hyperparameter settings.
At inference time, HyperRecon can efficiently produce diverse reconstructions, which would each correspond to different hyperparameter values.
In this framework, the user is empowered to select the most useful output(s) based on their own judgement. 
We demonstrate our method in compressed sensing, super-resolution and denoising tasks, using two large-scale and publicly-available MRI datasets.
Our code is available at \url{https://github.com/alanqrwang/hyperrecon}.
\end{abstract}


\section{Introduction}
The task of recovering high quality images from noisy or under-sampled measurements, often referred to as image reconstruction, is of crucial importance in many imaging applications.
Classically, image reconstruction is formulated as an ill-posed inverse problem and solved by optimizing an instance-based regularized regression loss function~\citep{fessler2019optimization}.
More recently, deep learning (DL), and in particular supervised learning, has shown promise in improving over classical methods due to its ability to learn from data and to perform fast inference~\citep{pal2021review}. 

The quality of solutions, for both the classical and DL-based techniques, depend heavily on the loss function that is minimized~\citep{ghodrati2019evaluation,zhao2018loss}. 
Different loss functions can highlight or suppress varying features or textures in the reconstructions.
A common approach is to minimize a composite loss that is a weighted sum of multiple terms. 
However, the tuning of the weight hyperparameters is a non-trivial problem, requiring methods such as grid-search, random search, or Bayesian optimization~\citep{frazier2018tutorial}.
Another weakness of existing methods is that once they are tuned and/or trained, they often produce a single best estimate of the reconstruction that is consistent with the measurements. 
%
Thus, any deviation from the conditions that the tool was optimized for means that the reconstructions can be sub-optimal.

More broadly, there is a lack of \textit{interactive} and \textit{controllable} tools that would enable human users to efficiently consider many reconstructions that are consistent with the measurements~\citep{Holzinger2016,xin2018accelerating}. 
Consider the results of the fastMRI Image Reconstruction Challenge\footnote{\url{https://ai.facebook.com/blog/results-of-the-first-fastmri-image-reconstruction-challenge}}, which revealed that commonly-used supervised metrics (e.g. mean-squared-error) do not correlate with the quality of images as judged by radiologists.
We argue that giving users control and choice over possible reconstructions is a human-in-the-loop approach to mitigate this discrepancy.  
%
\begin{figure*}[t!]
\begin{centering}
\includegraphics[width=\textwidth]{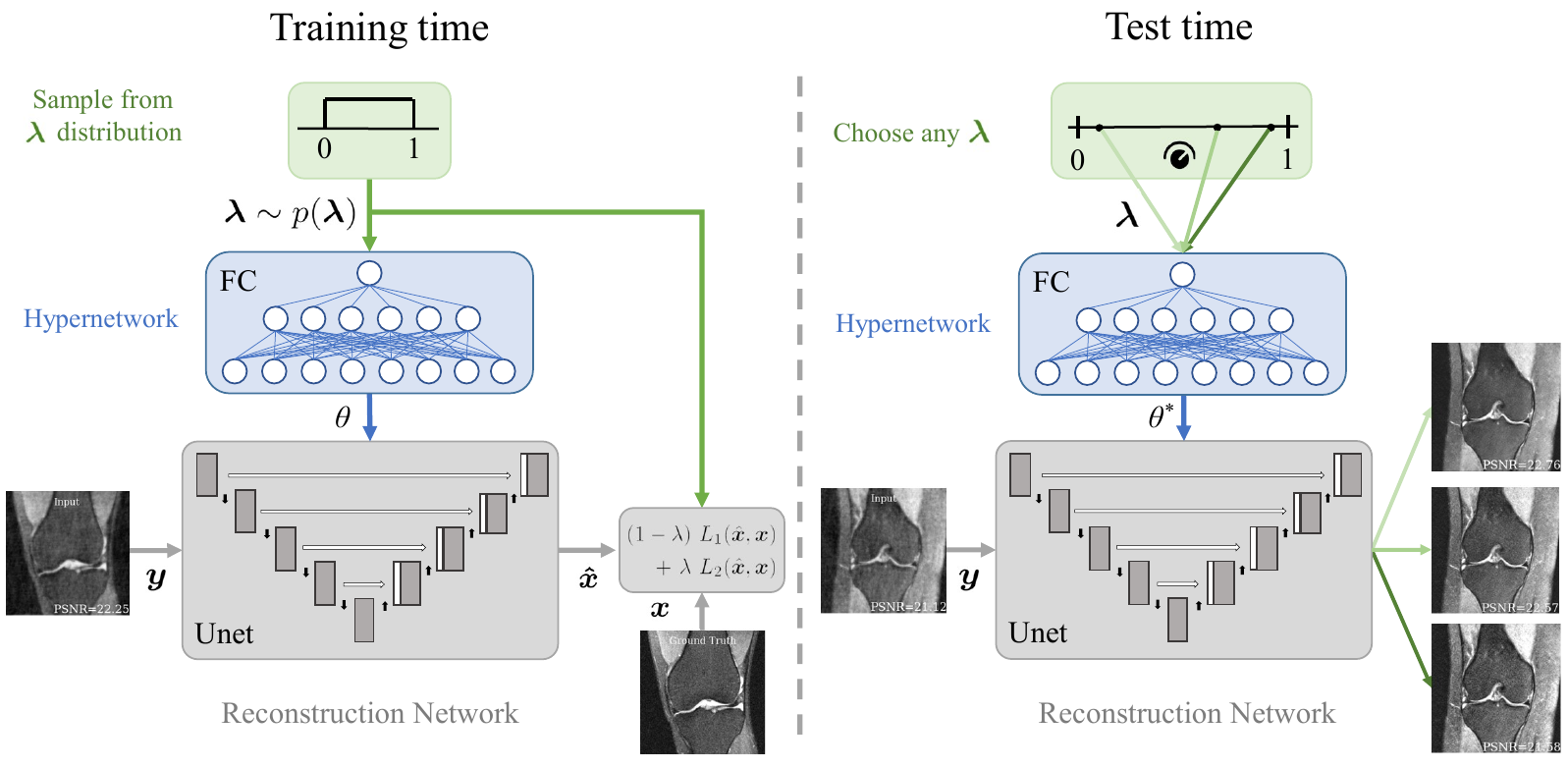}
\caption{Proposed model. Main reconstruction network (Unet) takes as input the under-sampled measurement $\bm{y}$ and outputs reconstruction $\bm{\hat{x}}$, while the fully-connected (FC) hypernetwork takes as input the hyperparameter $\bm{\lambda}$ and outputs weights $\theta$ for the reconstruction network. During training, $\boldsymbol{\lambda}$ is sampled from a uniform distribution. At test-time, $\bm{\lambda}$ can be interactively selected to efficiently compute reconstructions associated with that $\bm{\lambda}$, yielding diverse reconstructions.}
\label{fig:model-arch}
\end{centering}
\end{figure*}

In this work, to these ends, we propose a reconstruction strategy that is agnostic to the hyperparameter values and can efficiently produce reconstructions for a range of settings.
Specifically, our model uses a hypernetwork that takes as input hyperparameter value(s) and outputs the weights of the reconstruction network. 
At test-time, arbitrary hyperparameter values can be provided by the user and corresponding reconstructions will be efficiently computed via a forward-pass through the reconstruction network. 
Thus, our method is capable of producing a range of reconstructions corresponding to different hyperparameter values, instead of a single reconstruction associated with a single hyperparameter setting. 
The user can, in turn, interactively view and select from this set of possible solutions, for use in downstream tasks. 


Our method is applicable to any loss function where multiple terms need to be weighed. 
To demonstrate this, we perform experiments on two classes of loss functions in the reconstruction setting: supervised learning and amortized optimization. 
We experiment on three reconstruction tasks (CS-MRI, additive white Gaussian noise (AWGN) image denoising, and image superresolution) using two large-scale publicly available datasets (brain and knee MRI scans). These experiments highlight our proposed model's ability to match the performance of multiple baseline models trained on single settings of the hyperparameters.

This paper builds on our prior conference publication~\citep{wang2021hyperrecon}, which dealt with the unsupervised setting where only under-sampled data were available for training.
The present paper further considers the supervised setting, where we assume that we have access to fully-sampled training data and the primary design choice we are trying to solve is the weights in a composite loss function.
Our contributions are as follows:
\begin{itemize}
    \item We introduce a method for efficiently generating a range of image reconstructions using hypernetworks. Our method enables the end user to interactively view and choose the most useful reconstruction at inference time.
    \item We analyze several issues and propose solutions to hypernetwork training in the reconstruction setting, including hyperparameter distribution sampling and loss output scaling. Empirically, these solutions lead to improvement in model performance.
    \item Using multiple datasets and a variety of reconstruction tasks, we empirically demonstrate the performance of our proposed model and various improvements on two classes of loss functions: supervised learning (SL) and amortized optimization (AO). 
\end{itemize}

\section{Background}
\subsection{Inverse Problems for Image Reconstruction}
We assume an unobserved (vectorized) image $\boldsymbol{x} \in \mathbb{C}^N$ is transformed by a forward model $A$ into a measurement vector $\boldsymbol{y}$:
\begin{equation}
    \bm{y} = A\bm{x} + \epsilon.
    \label{eq:forward}
\end{equation}
Here,~$N$ is the number of pixels of the full-resolution grid and $\epsilon$ encapsulates unmodeled effects such as noise. 
In image denoising, $A = I$.
In image superresolution, $A = D$, where $D$ is a down-sampling operator.
In single-coil CS-MRI, $A = \mathcal{F}_u$, where $\mathcal{F}_u \in \mathbb{C}^{M \times N}$ is the under-sampled discrete Fourier operator and $M<N$.

Classically, inverse problems for image reconstruction are reduced to iteratively minimizing a regularized regression cost function on each collected measurement set~\citep{fista,admmboyd,primaldualchambolle,combettes2009proximal,daubechies2003iterative,Ye2019}. 
\begin{equation}
     \argmin_{\bm{x}} J(\bm{x}, \bm{y}) + \sum_{i=1}^{K-1} \lambda_i \mathcal{R}_i(\bm{x})  
     \label{eq:classical}
\end{equation}
The first term, called the data-consistency loss, quantifies the agreement between the measurement vector $\boldsymbol{y}$ and reconstruction. One common choice for $J$ is:
\begin{equation}
    J(\boldsymbol{x}, \boldsymbol{y}) = \norm{A \boldsymbol{x} - \boldsymbol{y}}^2_2.
    \label{eq:dcloss}
\end{equation}
The $K-1$ regularization terms $\mathcal{R}_i(\cdot)$ are hand-crafted priors which restrict the space of permissible solutions. A set of hyperparameters $\{\lambda_1,..., \lambda_{K-1}\}$ weights the competing contributions of the $K$ terms.
Common choices for these priors include sparsity-inducing norms of wavelet coefficients~\citep{waveletnorm} and total variation (TV)~\citep{liu2018image,hdtv,1992PhyD60259R}. 
While considerable research has been dedicated to designing suitable priors, such methods are invariably limited by lack of flexibility and adaptivity to the data. In addition, solving the minimization involves expensive iterative procedures that can take minutes per instance.

\subsection{Supervised Learning and Amortized Optimization}
Recent advances in data-driven algorithms and deep learning seek to address both aforementioned shortcomings of instance-based algorithms. Using large datasets and highly-parametrized and nonlinear neural network models, these algorithms learn priors directly from data and enable efficient inference via nearly-instantaneous forward passes.  

Given a training dataset $\mathcal{D}_{tr}$ and a neural network model $M_\theta$ with parameters $\theta$, the objective to minimize is:
\begin{equation}
    \argmin_\theta \mathbb{E}_{\mathcal{D}_{tr}}  \mathcal{L}(M_\theta; \bm{\lambda}),
\end{equation}
where $\mathcal{L}$ is a loss function parameterized by $\theta$, $\mathbb{E}_{\mathcal{D}_{tr}}$ denotes the empirical average, and $\bm{\lambda}$ encapsulates hyperparameters of interest.

In supervised learning (SL), $\mathcal{D}_{tr}$ is assumed to contain pairs of observations and corresponding ground-truth images $(\bm{y}, \bm{x}) \sim \mathcal{D}_{tr}$, and $\mathcal{L}$ is some combination of $K$ supervised losses $L(\cdot, \cdot)$: 
\begin{equation}
    \mathcal{L}(M_\theta; \bm{\lambda}) = L_K(M_\theta(\bm{y}),\bm{x}) + \sum_{k=1}^{K-1} \lambda_i L_i(M_\theta(\bm{y}),\bm{x}).
    \label{eq:supervised}
\end{equation}
Common choices for $L(\cdot, \cdot)$ include mean-squared error (MSE), mean-absolute error (MAE), and structural similarity index measure (SSIM). 

In addition to the SL set-up, we can consider the amortized optimization (AO) scenario, where $\mathcal{D}_{tr}$ is assumed to only contain observations $\bm{y} \sim \mathcal{D}_{tr}$. In AO, $\mathcal{L}$ takes a similar form to Eq.~\eqref{eq:classical}: 
\begin{equation}
    \mathcal{L}(M_\theta; \bm{\lambda}) = J(M_\theta(\bm{y}), \bm{y}) + \sum_{i=1}^{K-1} \lambda_i \mathcal{R}_i(M_\theta(\bm{y})).
    \label{eq:amortized}
\end{equation}
Thus, $M_\theta$ is trained to minimize Eq.~\eqref{eq:classical} for all observations in the dataset.
Although it minimizes the same cost function, AO provides several advantages over classical solutions. 
First, at test-time, it replaces an expensive iterative optimization procedure with a simple forward pass of a neural network.  
Second, since the model $M_\theta$ estimates the reconstruction for any viable input measurement vector $\bm{y}$ and not just a single instance, AO acts as a natural regularizer for the optimization problem~\citep{balakrishnan2019voxelmorph,shu2018amortized,wang2020amortized}. Third, recent work has shown that inductive biases of convolutional neural networks provide favorable implicit priors for image reconstruction~\citep{heckel2019deep,liu2018image,ulyanov2020dip}.
Since no ground-truth images are required, AO is sometimes referred to as unsupervised learning.

Importantly, similar to the classical cost function, both the SL and AO class of loss functions involve the tuning of a set of hyperparameters $\{\lambda_1,..., \lambda_{K-1}\}$, which can significantly affect the end reconstructions. 
Therefore, hyperparameter tuning is typically carefully done using expensive methods like cross-validation. In this work, we propose a strategy \textit{that can replace the expensive hyperparameter tuning step, such that reconstructions with different $\bm{\lambda}$ values can be efficiently generated at inference time.} To achieve this, we employ hypernetworks.

\subsection{Hypernetworks}
A hypernetwork is a neural network that generates the weights of a network which solves the main task (e.g. classification, segmentation, reconstruction).
In this framework, the parameters of the hypernetwork, and not the main network, are learned.
While originally introduced for achieving weight-sharing and model compression \citep{ha2016hypernetworks}, this idea has found numerous applications including neural architecture search \citep{brock2018smash,zhang2019graph}, Bayesian neural networks \citep{krueger2018bayesian,pmlr-v95-ukai18a}, multi-task learning \citep{lin2021controllable,mahabadi2021parameterefficient,pan2018hyperstnet,shen2017meta,Klocek2019}, and hyperparameter optimization~\citep{lorraine2018stochastic,hoopes2021hypermorph}. 

Notably, hypernetworks have been used to replace expensive hyperparameter tuning procedures like cross-validation~\citep{lorraine2018stochastic}. 
Hyperparameter tuning can be formulated as a nested optimization. The inner optimization minimizes some loss with respect to the weights of the main network $\theta \in \Theta$ over a training dataset $\mathcal{D}_{tr}$, while the outer optimization minimizes that loss with respect to the hyperparameters $\bm{\lambda} \in \Lambda$ over a held-out validation dataset $\mathcal{D}_{val}$: 
\begin{align}
\begin{split}
    \argmin_{\bm{\lambda}} \ \mathbb{E}_{\mathcal{D}_{val}}\mathcal{L} \left(\argmin_\theta \mathbb{E}_{\mathcal{D}_{tr}}{\mathcal{L}}(M_\theta; \bm{\lambda})\right) 
     := \argmin_{\bm{\lambda}} \ \mathbb{E}_{\mathcal{D}_{val}}{\mathcal{L}}\left( \theta^*(\bm{\lambda})\right).
\end{split}
\end{align}

Hypernetworks can be trained to approximate the solution to the inner optimization $\theta^*(\bm{\lambda})$. 
In this setting, a hypernetwork maps from hyperparameters to main network weights $H_\phi: \Lambda \rightarrow \Theta$. 
The model is trained via a stochastic optimization scheme whereby hyperparameters are sampled from a pre-defined distribution over $\Lambda$ during training:
\begin{equation}
    \phi^* = \argmin_\phi \mathbb{E}_{\Lambda} \mathbb{E}_{\mathcal{D}_{tr}}\mathcal{L}(M_{H_\phi(\bm{\lambda})}; \bm{\lambda}).
    \label{eq:hypernet-objective}
\end{equation}
The approximate solution to the inner optimization is then $\theta^*(\bm{\lambda}) = H_{\phi^*}(\bm{\lambda})$.
Note that only the parameters of the hypernetwork $\phi$ are learned. 

The upshot of this method is that rapid hyperparameter tuning is possible, since the inner optimization can be performed with a forward pass of the hypernetwork. While using this model enables us to address the issue of costly hyperparameter tuning in the context of image reconstruction (as discussed in the previous section), we further build on this idea in this work by \textit{leveraging the vast set of reconstructions that are capable of being generated for a given measurement as a result of changing $\bm{\lambda}$.} That is, we propose this method as a means of creating an interactive and controllable image reconstruction tool.

\section{Related Works}
Hypernetworks have recently shown promise in rendering agnosticm to hyperparameters with minimal increase in training time. 
Hoopes et. al. use this method to enable test-time rapid tunability in image registration for medical imaging~\citep{hoopes2021hypermorph}.
Our previous conference work applied this idea to the AO setting for CS-MRI~\citep{wang2021hyperrecon}. In this work, we extend this idea broadly to other image reconstruction tasks in both the SL and AO settings, and validate on additional datasets.

Sahu et. al. propose an interactive method for optimizing smoothing parameters in Digital Breast Tomosynthesis~\citep{sahu2021interactive}. The authors' method produce reconstructions conditioned on a user-specified smoothing parameter by multiplying the intermediate activations of a reconstruction network with the parameter; the network is trained to minimize mean absolute error. However, this method requires ground-truth reconstructions obtained by solving many instance-based iterative procedures, which is computationally expensive.

Our work is similar in spirit to the work of Lin et. al., which uses hypernetworks to enable controllability in the multi-task learning setting~\citep{lin2021controllable}. 
In their work, they pose multi-task learning as a multi-objective minimization problem, and show that hypernetworks can enable test-time trade-off control among different tasks (e.g. depth prediction and semantic segmentation) with a single model. 

\section{Methods}
We propose to use hypernetworks for image reconstruction, where our goal is to produce a model that can efficiently compute a reconstruction that approximates the solution to Eqs.~\eqref{eq:supervised} and~\eqref{eq:amortized} for arbitrary hyperparameter coefficient values.
We call this model a \textit{controllable} reconstruction network, since the resulting model can produce a diverse set of reconstructions that can be controlled interactively at test-time. We illustrate the model in Fig.~\ref{fig:model-arch}.

\subsection{Controllable Reconstruction Network}
Let $M_\theta$ denote a main network with parameters $\theta \in \Theta$ which maps an observation $\bm{y}$ to a reconstruction $\hat{\bm{x}}$. 
We define a hypernetwork $H_\phi : \mathbb{R}_+^{K-1} \rightarrow \Theta$ that maps a hyperparameter weight vector $\boldsymbol{\lambda}$ to the parameters $\theta$ of the main network $M_\theta$. A reconstruction for a given observation $\bm{y}$ and hyperparameter vector $\bm{\lambda}$ is then $\hat{\bm{x}} = M_{H_\phi(\bm{\lambda})}(\bm{y})$. Effectively, this makes $\bm{\lambda}$ an input to the model.

The objective is given in Eq.~\eqref{eq:hypernet-objective}, where $\Lambda = \mathbb{R}_+^{K-1}$.

\subsection{Training}
We restrict our attention to the case of two and three loss terms (one and two hyperparameters, respectively), although we emphasize that our method is applicable to an arbitrary number of loss terms. 
In general, Eqs.~\eqref{eq:supervised} and \eqref{eq:amortized} can be manipulated such that the hyperparameter support is bounded to $\bm{\lambda} \in \Lambda = [0,1]^{K-1}$. 
For example, for one and two hyperparameter weights and arbitrary loss terms $L_1, L_2$, and $L_3$ (omitting $\theta$, $\bm{x}$, and $\bm{y}$ for brevity):
\begin{align}
\label{eq:one-hyperparameter}
\begin{split}
    \mathcal{L}_{2} &= (1-\lambda)L_1 + 
    \lambda L_2,
\end{split} \\
\label{eq:two-hyperparameter}
\begin{split}
 \mathcal{L}_{3} &= \lambda_1 L_1 + 
    (1-\lambda_1)\lambda_2 \ L_2 + (1-\lambda_1)(1-\lambda_2)\ L_3. 
\end{split}
\end{align}
\begin{wrapfigure}{r}{0.5\textwidth}
\begin{minipage}{0.5\textwidth}
\begin{algorithm}[H]
    \caption{UHS and DHS Hypernetwork Training}\label{algorithm1}
    Input: Dataset $\mathcal{D}_{tr}$, model $M_{H_\phi}$, $K$, $B$, $b < B$, $\gamma>0$ \\
    Output: Model weights $\phi$
    \begin{algorithmic}[1]
    \Repeat
        \State Sample $\{\bm{y}_1,..., \bm{y}_B\} \sim \mathcal{D}_{tr}$ 
        \State Sample $\{\bm{\lambda}_1, ..., \bm{\lambda}_B\} \sim U[0,1]^{K-1}$
        \State $\hat{\bm{x}}_i \leftarrow M_{H_\phi(\bm{\lambda}_i)}(\bm{y}_i)$ for $i = 1,...,B$
        \If{UHS}
            \State $\phi \leftarrow \phi - \gamma \nabla_\phi\sum_{i=1}^{\textcolor{red}{B}} \mathcal{L}_K\left(\hat{\bm{x}}_i; \textcolor{red}{\bm{\lambda}_i}\right)$
        \ElsIf{DHS}
            \State $\{\tilde{\bm{\lambda}}_1, ..., \tilde{\bm{\lambda}}_B\} \leftarrow \text{sort}\{\bm{\lambda}_1, ..., \bm{\lambda}_B\}$ 
            \State $\phi \leftarrow \phi - \gamma \nabla_\phi\sum_{i=1}^{\textcolor{red}{b}} \mathcal{L}_K\left(\hat{\bm{x}}_i; \textcolor{red}{\tilde{\bm{\lambda}}_i}\right)$
        \EndIf
    \Until{convergence} \\
    \Return $\phi$
    \end{algorithmic}
\end{algorithm}
\end{minipage}
\hfill
\caption{Algorithm for UHS and DHS training. Subscripts index over the mini-batch. Sorting (Line 8) is done in ascending order w.r.t. $J\left(\hat{\bm{x}}_i, \bm{y}_i\right)$. Red highlights differences between UHS and DHS.}
\end{wrapfigure}

A straightforward strategy for training the hypernetwork involves sampling the coefficients from a uniform distribution $p(\bm{\lambda}) = U[0,1]^{K-1}$ and sampling an under-sampled measurement vector $\bm{y}$ for each forward pass during training. 
The gradients are then computed with respect to the loss evaluated at the sampled $\bm{\lambda}$ via a backward pass.
This corresponds to minimizing Eq.~\eqref{eq:hypernet-objective} with a uniform distribution for~$\bm{\lambda}$.
We denote this sampling strategy as uniform hyperparameter sampling (UHS).

\subsection{Data-driven Hyperparameter Sampling for AO}

In the hypothetical scenario of infinite hypernetwork capacity, the hypernetwork can capture a mapping of any input $\bm{\lambda}$ to the optimal $\theta^* = H_{\phi^*}(\bm{\lambda})$ that minimizes Eqs.~\eqref{eq:supervised} and~\eqref{eq:amortized} with the corresponding $\bm{\lambda}$. 
However, in practice, the finite model capacity of the hypernetwork constrains the ability to achieve optimal loss for every hyperparameter value. 
In this case, the model performance will depend on the adopted distribution for $\bm{\lambda}$. 

In SL loss functions, we are typically interested in the reconstructions associated with all possible settings of the coefficients. Thus, UHS is a necessary sampling strategy.
In contrast, in AO loss functions, some coefficients will not produce acceptable reconstructions, even if solved optimally. 
For example, a reconstruction with a large coefficient for TV tends to be overly smooth with little to no structural content.
Thus, sampling hyperparameters from the entire support $[0,1]^{K-1}$ ``wastes" model capacity on undesirable regions of hyperparameter space. 

In most real-world reconstruction scenarios, we don't have a good prior distribution from which to sample desirable regions. 
Instead, we propose a data-driven sampling scheme (DHS) which learns the prior over the course of training. 
We leverage the data-consistency loss induced by a setting of the coefficients to assess whether the reconstruction will be useful or not. 
Intuitively, values of $\bm{\lambda}$ which lead to high data-consistency loss $J\left(M_{H_\phi(\bm{\lambda})}(\bm{y}), \bm{y}\right)$ will produce reconstructions that deviate too much from the underlying anatomy, and which therefore can be ignored during training.

One can enforce this idea by computing gradients only on values of $\bm{\lambda}$ which induce a data-consistency loss below a pre-determined threshold during training. 
However, this presents two problems. 
First, calibrating and tuning this threshold can be difficult in practice. 
Second, at the beginning of training, since the main network will not produce good reconstructions, this threshold will likely not be satisfied. 

In lieu of a complex training/threshold scheduler, we adapt the threshold with the quality of reconstructions directly within the training loop.
The proposed DHS strategy works by using the best $b$ samples with the lowest data-consistency loss within a mini-batch of size $B$ to compute the gradients. 
In effect, this induces a variable threshold $b/B$ of the landscape percentage which is optimized, where this threshold adapts dynamically over training. 
Algorithm~\ref{algorithm1} details the training loop for both UHS and DHS. 

We demonstrate in the Experiments section that DHS leads to improved model performance on the most promising hyperparameter space regions compared to UHS, given a fixed hypernetwork capacity.

\subsection{Matched Loss Output Scales}
\label{sec:matched}
In general, a given loss function $L_i$ can be arbitrarily scaled by a constant $\alpha_i L_i$ and still yield the same reconstruction results (ignoring the influence on learning rate, for example).
In multi-term loss functions, the relative scales of the $K$ loss outputs should be approximately matched during training to allow for equal contribution of gradients during backpropagation. 

In the context of our proposed method, these loss output scales have an added effect: different scales will change the resulting reconstructions as a function of $\bm{\lambda}$. 
If one loss dominates another, then most of the landscape will be dedicated to minimizing the dominant loss and very little variation will exist for different $\bm{\lambda}$.
Indeed, we ideally would like a landscape which varies maximally over $\bm{\lambda}$, which would (hopefully) capture as wide a range of diverse reconstructions as possible. 

For SL, we account for this by normalizing the loss functions by its ``best-case" loss averaged over a validation set. 
Let $s_i$ denote the average validation loss on $L_i$ for a model only optimized for this specific loss function, and we set $\alpha_i = 1/s_i$.
During training, validation losses associated with all loss terms will (roughly) converge to a value of $1$, and thus the loss output scales will be approximately matched. 
Further details are given in Section~\ref{sec:loss_functions}. 

For AO, we cannot take the same approach since the ``best-case" losses for $J(\cdot, \cdot)$ and $\mathcal{R}(\cdot)$ are $0$. 
When Eq.~\eqref{eq:classical} is derived from the maximum a posteriori (MAP) estimate, the scaling factors can be computed with respect to the parameters of the likelihood and/or prior distributions~\citep{chambolle2010intrototv,hoopes2021hypermorph}. 
However, in cases where the regularization term cannot be expressed as a valid probability or when multiple terms are used, such an approximation cannot be obtained.
In this work, we tuned the scaling factors to align the magnitudes of the separate terms. 
Although this works well in practice, this issue is still an open question in the general case and may be an important direction for further research.

\section{Experiments}
\label{sec:experiments}


We evaluate our proposed method on two large, publicly-available MRI datasets consisting of coronal knees and axial brains for SL and AO classes of multi-term loss functions. For AO, we experiment with the CS-MRI reconstruction task. For SL, we experiment with CS-MRI, AWGN denoising, and image superresolution reconstruction tasks.

\subsection{Data}
We conducted our experiments using two large, publicly-available datasets. The first is the NYU fastMRI dataset~\citep{zbontar2018fastmri} composed of proton density (PD) and proton density fat-suppressed (PDFS) weighted knee MRI scans.
The second is the ABIDE dataset, composed of T1-weighted brain MRI scans~\citep{abide}.

For both datasets, we used 100, 25, and 50 subjects for training, validation, and testing, respectively.
Volumes were separated into 2D slices and slices with only background were removed, resulting in a final train/val/test split of 3500/875/1750 for knee and 11400/2850/5700 for brain slices. 
All images were intensity-normalized to the range $[0,1]$ and cropped and re-sampled to a pixel
grid of size $256 \times 256$.

\subsection{Model Details}
This work presents a general strategy for computing hyperparameter-agnostic reconstructions with a single model, which is applicable to any main network architecture.
In this work, the main network $M_\theta$ is a commonly-used Unet architecture~\citep{unet} adapted for reconstruction, where the network input is a 2-channel, complex-valued input and the network output is a single-channel, real-valued output.
A hidden channel dimension of $32$ is used for all layers, resulting in a total main network parameter count of $149,409$.

Since $M_\theta$ receives noisy images as input, this network can be viewed as a post-processing network preceded by an optional model-based inversion step (e.g. applying the inverse Fourier transform to under-sampled measurements), which is a common reconstruction pipeline in the literature~\citep{jin2017deepconvolutional,kang2017lowdose,wang2021jointoptimization}.
We expect the conclusions drawn in this work to hold for any deep learning-based main network, particularly state-of-the-art unrolled networks~\citep{monga2019algorithmunrolling}.

The hypernetwork $H_\phi$ consists of a 5-layer fully-connected (FC) architecture with intermediate LeakyReLUs. The input to the network is the number of hyperparameters, and the hidden dimension and output dimension are of size $d$. 
Each convolution layer in the main network takes as input the $d$-dimensional output embedding of the hypernetwork and linearly projects it to the size of the kernel and bias, where the parameters of the projection are learned. 
We treat the hypernetwork and the main network as a single, large network, and refer to the overall model as HyperRecon.
We experiment with hypernetwork hidden dimensions $d \in \{4, 32, 128\}$, and refer to them as HyperRecon-\{S, M, L\}, respectively.
We used a batch size of 32 and the Adam optimizer for all models~\citep{kingma2017adam}.

All training and testing experiments were performed on a machine equipped with an Intel Xeon Gold 6126 processor and an NVIDIA Titan Xp GPU. All models were implemented in Pytorch. Table~\ref{tab:times} outlines the training time, inference time, and number of parameters for all models used in experiments. We train each model until the loss converges on the validation set, typically for 1000 epochs.

\subsection{Baselines} 
For comparison, we trained separate Unet reconstruction networks for each fixed hyperparameter value. 
We refer to these models as baselines and emphasize that they demand significant computational resources, since each of these models needs to be trained and saved separately (see Table 1). 
For SL experiments, we trained five baseline models for $\lambda \in \{0.0, 0.25, 0.5, 0.75, 1.0\}$. 
For AO experiments, we trained $361$ baseline models with hyperparameters chosen non-uniformly on a $19\times 19$ grid over the space $[0,1]\times [0,1]$, in order to more densely sample in high-performing regions\footnote{ The 19 values were \{0.0,0.1,0.2,0.3,0.4,0.5,0.6,0.7,0.8,0.85,0.9,0.93,0.95,0.98, 0.99,0.995,0.999,1.0\}.}.

\begin{SCtable}
\centering \footnotesize
\begin{tabular}{@{}lcc@{}}
\toprule
\textbf{\textit{Model}}     & \textbf{\textit{Train time (hrs)}} & \textbf{\textit{\# parameters}} \\ \midrule
Unet      & $\sim$4                 & 149K\\
All Unets      & $\sim$648          & 53M\\ \midrule
HyperRecon-S & $\sim$5               & 744K\\ 
HyperRecon-M & $\sim$6               & 4.9M\\ 
HyperRecon-L & $\sim$7               & 19M\\ 
\bottomrule
\end{tabular}
\caption{Training time and number of parameters for all models on brain dataset. ``$\sim$" denotes approximation. ``All Unets" denotes all $361$ baselines across all hyperparameters. Inference time (defined as the runtime of one forward pass with a single $\bm{\lambda}$ and $\bm{y}$ input, averaged over the test set) for all models is $\sim$0.2 seconds.}
\label{tab:times}
\end{SCtable}

\subsection{Loss Functions}
\label{sec:loss_functions}
\subsubsection{Supervised Learning} 
\label{sec:loss_functions_sl}
For SL, we consider CS-MRI, denoising, and superresolution reconstruction tasks. For all tasks, we use MAE for $L_1$ and SSIM for $L_2$~\citep{ssim} as the two loss terms in Eq.~\eqref{eq:one-hyperparameter}.

MAE is a global, intensity-based metric which is simple and widely-used. SSIM is a perception-based metric that considers image degradation as perceived change in structural information. Prior work has shown performance gain when using a combination of these two losses~\citep{zhao2018loss}.
While we restrict our focus to these two losses in the SL setting, we emphasize that our method works for any choice of supervised loss functions. 

Proper HyperRecon training requires approximately-matched loss output scales, i.e. $\alpha_1$ and $\alpha_2$ (see Section~\ref{sec:matched}). To compute these values, we first train two Unet models for $\lambda=0$ and $\lambda=1$. The validation losses $s_1$ and $s_2$ for the two models are computed, and then we set $\alpha_1=1/s_1$ and $\alpha_2 = 1/s_2$ during HyperRecon training.

\subsubsection{Amortized Optimization} 
For AO, we consider the CS-MRI reconstruction task and experiment with two regularization terms $\mathcal{R}_1$ and $\mathcal{R}_2$ in Eq.~\eqref{eq:two-hyperparameter}:
layer-wise total $\ell_1$-penalty on the weights~$\theta$ of the main reconstruction network and the anisotropic total variation of the reconstruction image:
\begin{align}
\begin{split}
    \mathcal{R}_1 &= \sum_{i=1}^L \norm{\theta_i}_1, \\
\end{split}
\begin{split}
    \mathcal{R}_2 &= \sum_{i=1}^N \big|[D_1 x]_i\big| + \big|[D_2 x]_i\big|,
\end{split}
\end{align}
where $\theta_i$ denotes the weights of the reconstruction network for the $i$th layer, $L$ is the total number of layers of the reconstruction network, and $D_1$ and $D_2$ denote the finite difference operation along the first and second dimension of a two-dimensional image.

\subsection{Evaluation}
\subsubsection{Supervised Learning}
For SL experiments, in addition to evaluating on the trained losses MAE and SSIM, we also evaluate on PSNR and high-frequency error norm (HFEN)~\citep{ravishankar2011dictionary}.
Given a $15 \times 15$ Laplacian-of-Gaussian (LoG) filter with a standard deviation of $1.5$ pixels, the HFEN is computed as the $\ell_2$ difference between LoG-filtered ground truth and LoG-filtered reconstructions.

To demonstrate the downstream utility of HyperRecon, we evaluate segmentation performance of the reconstructions for varying $\lambda$ on a deep learning-based segmentation model. 
The model has the same architecture as the baseline Unet $M_\theta$, except that it has a 5-channel output and a final softmax layer. 
We train the model to minimize the soft-Dice loss~\citep{sudre2017generalizeddice} on the same ABIDE dataset with the same data split as reconstruction training, and evaluate the hard-Dice score ~\citep{lee1945dice}. Ground truth segmentation maps corresponding to 5 regions-of-interest\footnote{Background, gray matter, white matter, cerebrospinal fluid, and cortex.} were anatomically segmented with FreeSurfer~\citep{fischl2012freesurfer} (see~\citep{balakrishnan2019voxelmorph} for details).
Batch size, optimizer, and other hyperparameters were identical to reconstruction training.

\subsubsection{Amortized Optimization}
For AO experiments, we report the \textit{relative} PSNR (abbreviated rPSNR) for a reconstruction by subtracting the PSNR value for the zero-filled reconstruction from the PSNR value of the reconstruction. 
Positive values for rPSNR are preferable as they indicate that the regularization terms lead to improvement over the trivial zero-filled reconstructions.  

For ease of visualization, we report the \textit{negative} of certain metrics so that higher values are better for all metrics. These are abbreviated by an ``n" in front of the metric name (e.g. nMAE, nHFEN). 

\subsection{Reconstruction Tasks}
For all tasks, noisy inputs to our model were obtained retrospectively and performed \textit{in-silico}. 
Although this is a slight simplification, we believe that these experiments are consistent with the main methodological message of this work, and furthermore that these results will translate to real-world scenarios.
\subsubsection{CS-MRI}
$k$-space data was generated by retrospective under-sampling using
4-fold and 8-fold acceleration under-sampling masks generated using a polynomial Poisson-disk variable-density sampling strategy~\citep{Geethanath2013CompressedSM,lustig}. 
The input into the models were the zero-filled reconstructions, i.e. the inverse Fourier transform of the under-sampled $k$-space data with zeros for missing values.

\subsubsection{AWGN Denoising}
Ground truth images were used as clean images, which were corrupted with additive white Gaussian noise sampled from $\mathcal{N}(0, \sigma^2 I)$, where $\sigma=0.1$. 

\subsubsection{Superresolution}
Ground truth images were used as full-resolution images, which were down-sampled 
by a factor of $4$ and subsequently up-sampled using nearest-neighbor interpolation back to the original grid size.

\section{Results}
\subsection{Supervised Learning}
We evaluate the performance of HyperRecon-L using metric curves over the space of permissible hyperparameter values $\lambda \in [0,1]$. 
We generated the curves for visualization by densely sampling the support $[0,1]$ to create $100$ discrete samples.
For each grid point, we computed the value by passing the corresponding hyperparameter values to the model along with each under-sampled measurement $\bm{y}$ in the test set and taking the average PSNR value. 

\begin{figure*}[t]
\begin{centering}
\includegraphics[width=\textwidth]{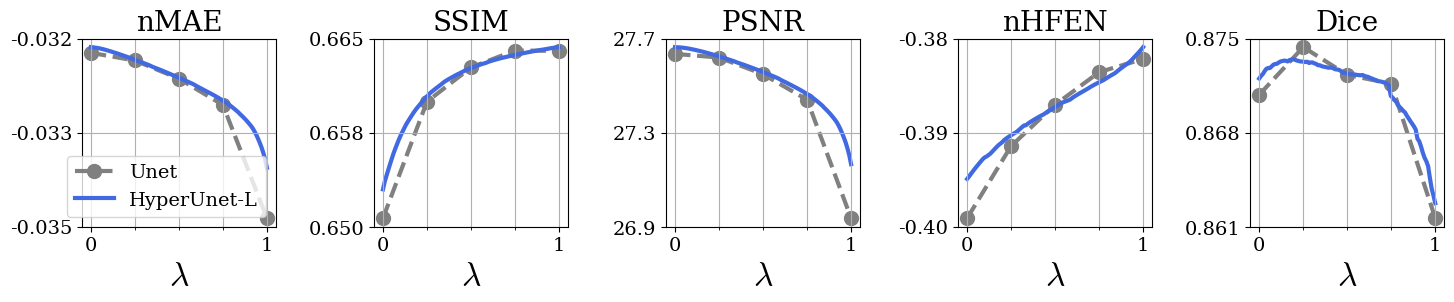}
\caption{Model performance on brain dataset for SL loss functions. Higher is better. Each panel shows metric performance for Unet and HyperRecon-L models for varying $\lambda$. The loss function of the HyperRecon-L model is a sum of MAE and SSIM loss functions (first 2 panels), where $\lambda=0$ corresponds to only MAE loss and $\lambda=1$ corresponds to only SSIM loss (see Section~\ref{sec:loss_functions_sl}).}
\label{fig:supcurves_brain}
\end{centering}
\end{figure*}

\begin{figure}[t]
\begin{centering}
\includegraphics[width=\textwidth]{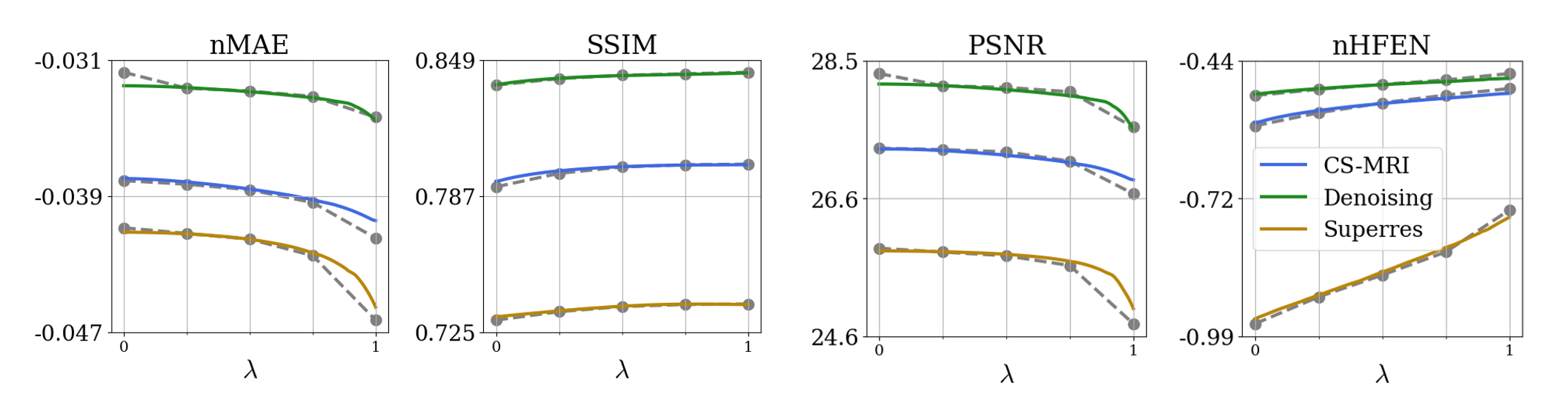}
\caption{Model performance on knee dataset for SL loss functions. Higher is better. Each panel shows metric performance for Unet and HyperRecon-L models for varying $\lambda$, where each color corresponds to a different reconstruction task. The loss function of the HyperRecon-L model is a sum of MAE and SSIM loss functions (first 2 panels), where $\lambda=0$ corresponds to only MAE loss and $\lambda=1$ corresponds to only SSIM loss (see Section~\ref{sec:loss_functions_sl}).}
\label{fig:supcurves_knee}
\end{centering}
\end{figure}

Fig.~\ref{fig:supcurves_brain} shows metric performance for brain data.
Each panel shows Unet and HyperRecon-L performance for the specified metric across $\lambda$ values. Similarly, Fig.~\ref{fig:supcurves_knee} shows metric performance for knee data on three different reconstruction tasks. 
We observe very close matching of performance across all metrics between Unet and HyperRecon-L models, indicating that the hypernetwork is able to generate the optimal weights for the entire hyperparameter space.
For downstream brain segmentation, the optimal $\lambda$ for maximizing Dice is around $0.25$.
With HyperRecon-L, this optimal point can be easily obtained with a single model.

\begin{figure}
\centering
\begin{subfigure}[b]{0.95\textwidth}
   \includegraphics[width=1\linewidth]{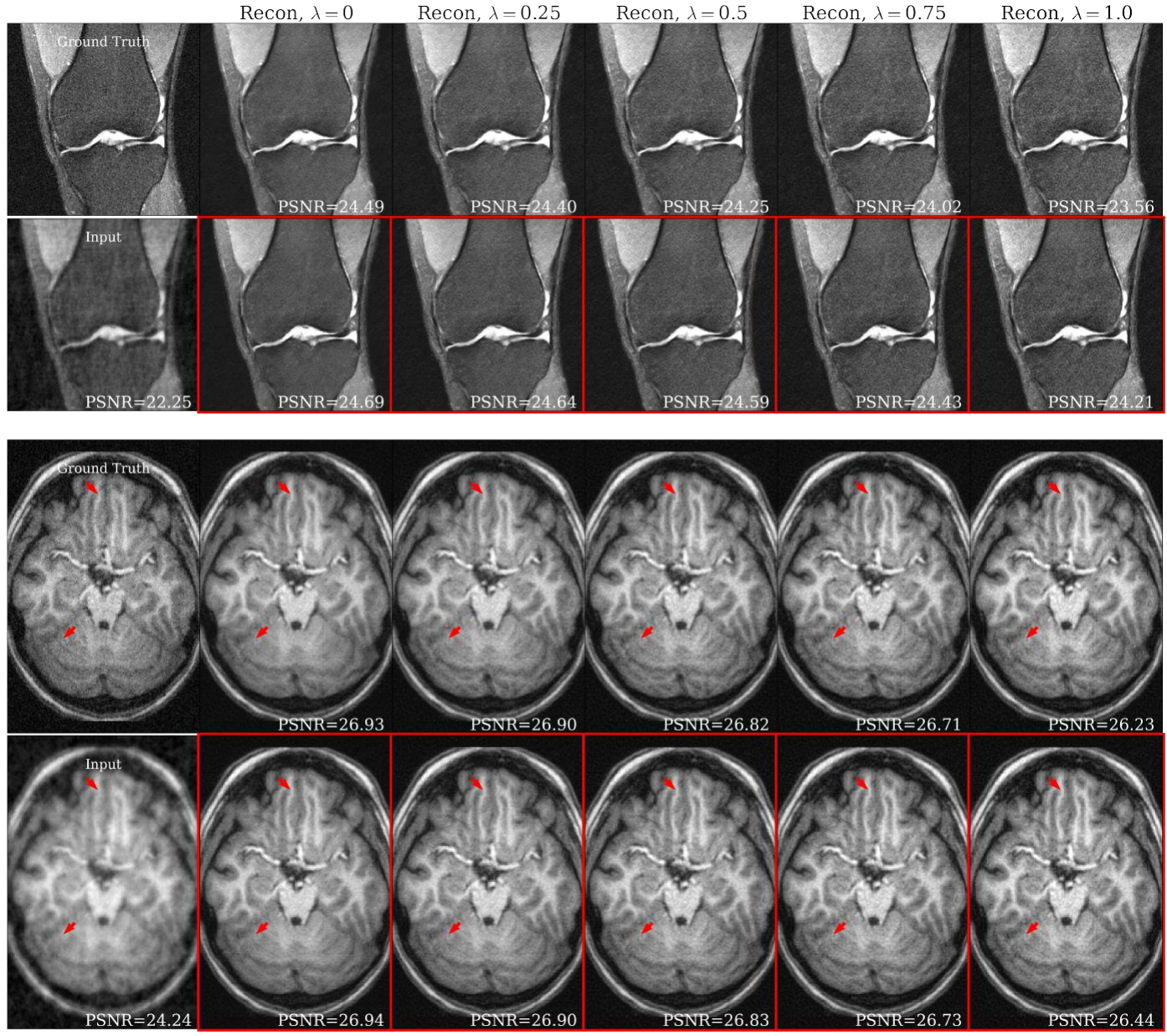}
   \caption{}
   \label{fig:main_sup} 
\end{subfigure}

\begin{subfigure}[b]{0.95\textwidth}
   \includegraphics[width=1\linewidth]{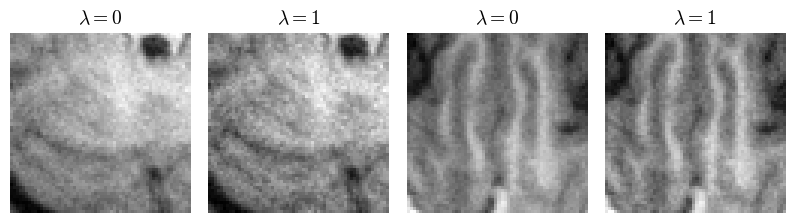}
   \caption{}
   \label{fig:main_sup_patches}
\end{subfigure}

\caption[Supervised representative slices.]{(a) Representative slices for SL on CS-MRI task with 8-fold under-sampling. Ground truth and input images are denoted in the first column. Top row are baseline Unet reconstructions with varying $\lambda$. Bottom row (red) are HyperRecon-L reconstructions with varying $\lambda$. PSNR values in reconstructions provide a basis of comparison between Unet and HyperRecon-L reconstructions of the same $\lambda$. Brain slices are cropped to show detail. Arrows indicate corresponding points where the difference between reconstructions can be appreciated. (b) Representative brain patches for HyperRecon-L, centered around red arrows.}
\end{figure}

Fig.~\ref{fig:main_sup} shows representative brain and knee slices for the CS-MRI task, and Fig.~\ref{fig:main_sup_patches} shows zoomed-in versions of the red arrow regions for the brain slices.
We notice a significant visual similarity between the same hyperparameters for Unet and HyperRecon-L models. 
In general, a higher weight on MAE tends to lead to more blurry and smoother reconstructions, while a higher weight on SSIM tends to lead to more noisy reconstructions with more high frequency content. In addition, MAE tends toward less contrast between hypo-intense and hyper-intense regions, whereas SSIM accentuates this contrast more.
With baseline models where $\bm{\lambda}$ must be chosen before training, these variations would be completely missed and end users would be stuck with one reconstruction. 
Additional slices for CS-MRI, denoising, and superresolution tasks are presented in the Appendix.

%
%
%
\begin{figure}[t]
\begin{centering}
\includegraphics[width=\textwidth]{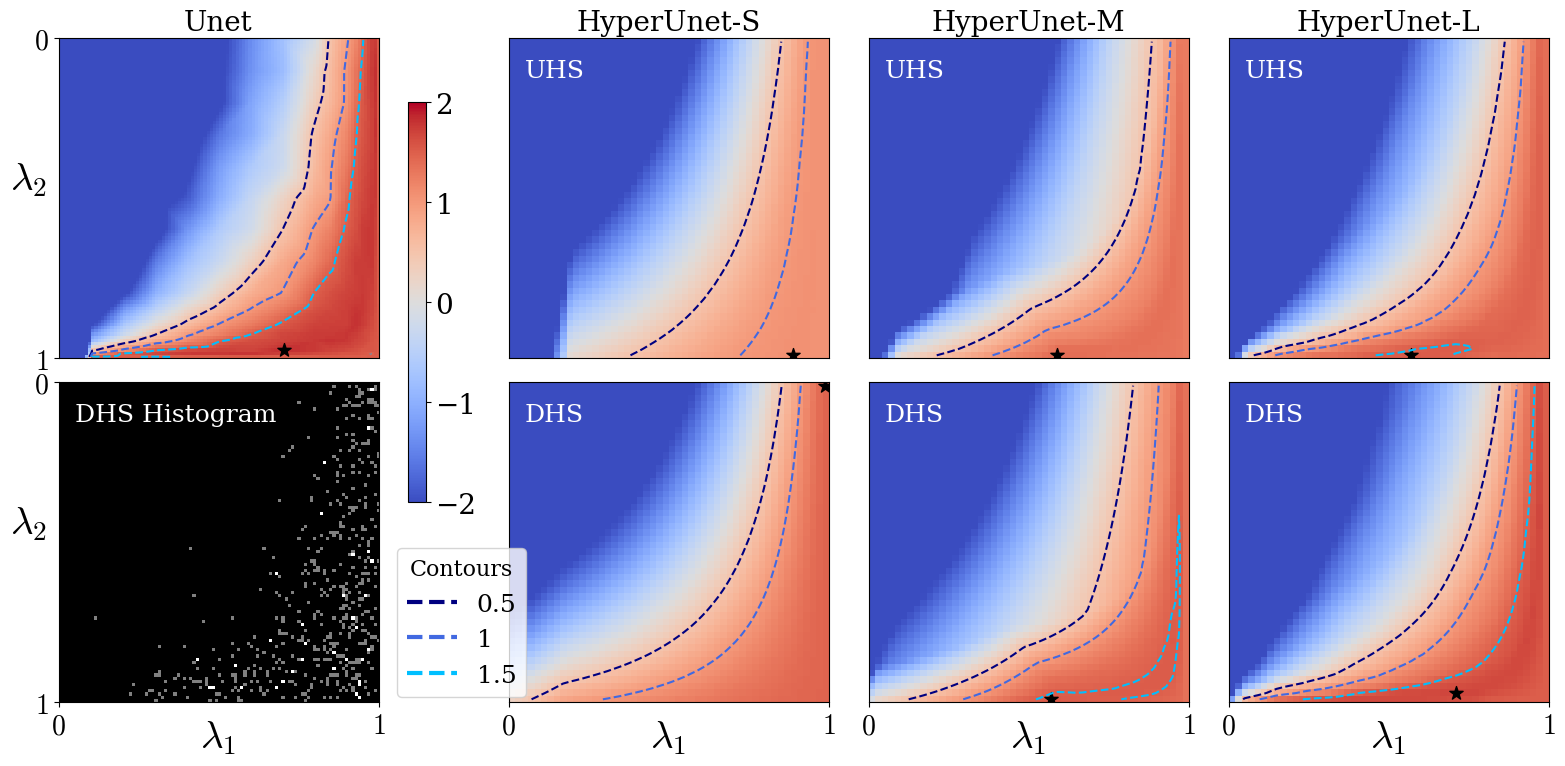}
\caption{rPSNR values for knee dataset on AO loss over the hyperparameter support $[0,1]\times [0,1]$ for different hypernetwork capacity and sampling methods. 
The $x$-axis and $y$-axis denote the value of the hyperparameters $\lambda_1$ and $\lambda_2$, respectively. Contours denote level sets of fixed value (see legend). Stars in the landscapes denote maximum value. (Left) Top image depicts the baseline landscape. Bottom image shows an example histogram of hyperparameter values used for gradient computation during one epoch of training with the DHS strategy. (Right) Top and bottom row show the UHS and DHS model landscapes, respectively, for three different hypernetwork capacities.}
\label{fig:landscapes}
\end{centering}
\end{figure}
\begin{figure*}[t!]
\begin{centering}
\includegraphics[width=\textwidth]{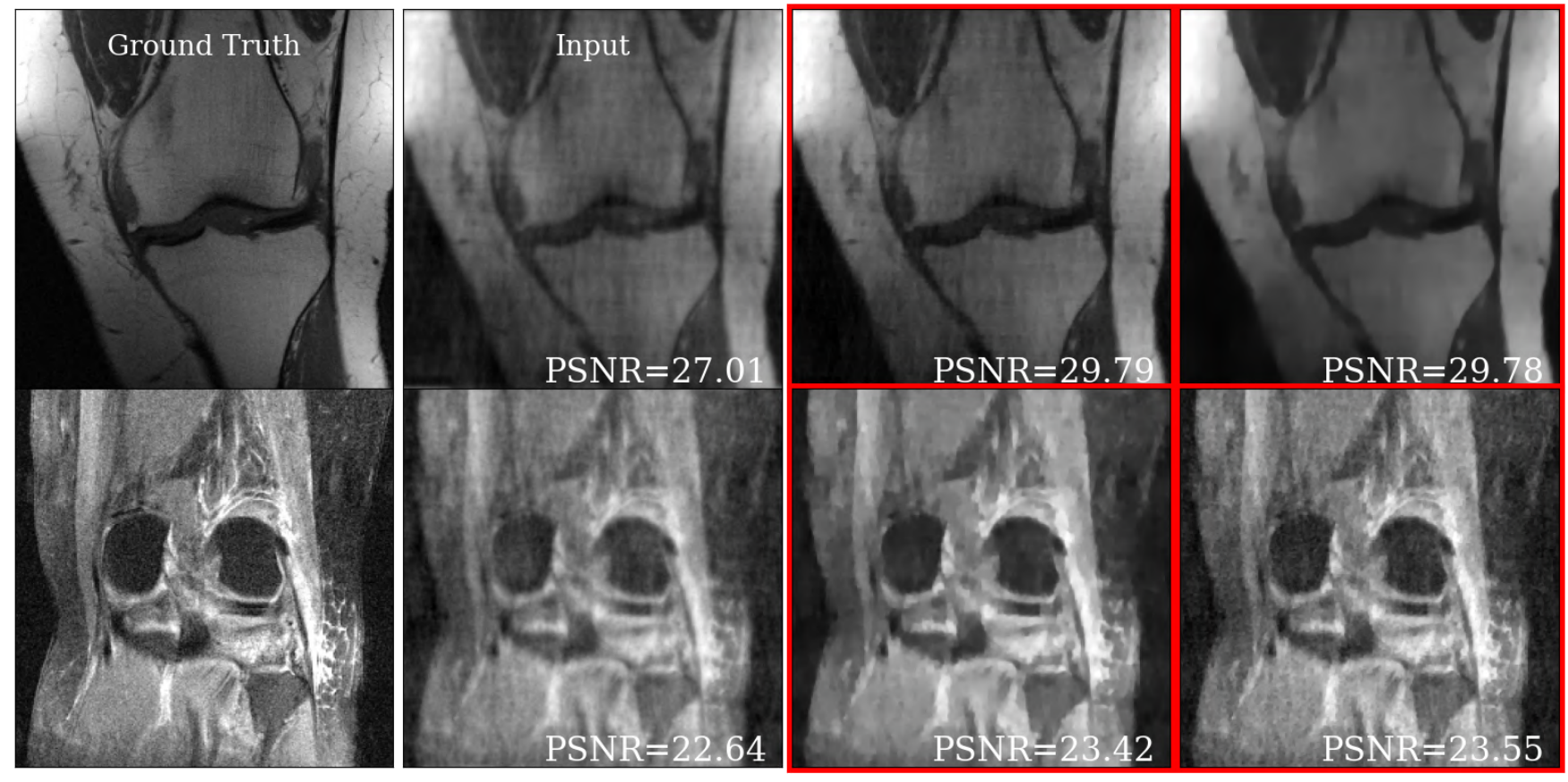}
\caption{Representative slices for AO on CS-MRI task on 8-fold undersampled knees. Ground truth is denoted and two visually different but high-quality reconstructions (in terms of PSNR) are shown in red.}  
\label{fig:main_ao}
\end{centering}
\end{figure*}

\subsection{Amortized Optimization}
We evaluate the performance of HyperRecon-\{S, M, L\} using rPSNR landscapes over the space of permissible hyperparameter values $(\lambda_1, \lambda_2) \in [0,1]\times [0, 1]$. 
We generated landscapes for visualization by densely sampling the support $[0,1]\times [0,1]$ to create a grid of size $100 \times 100$.
For baselines, the $19 \times 19$ grid was linearly interpolated to $100 \times 100$ to match the hypernetwork landscapes.

Fig.~\ref{fig:landscapes} shows rPSNR landscapes for different reconstruction models. 
The top left map corresponds to baselines. 
The first image in the second row shows an example histogram of the regularization weight samples used for gradient computation over one DHS training epoch.  

The remaining images in the top row correspond to UHS models with varying hypernetwork capacities. 
Similarly, the bottom row shows rPSNR landscapes for DHS models at the same capacities. 
We find that higher capacity hypernetworks approach the baseline models' performance, at the cost of computational resources and training time (see Table~\ref{tab:times}). 
We also observe significant improvement in performance using DHS as compared to UHS, given a fixed hypernetwork capacity. 
We find that the performance improvement achieved by DHS is less for the large hypernetwork, validating the expectation that the sampling distribution plays a more important role when the hypernetwork capacity is restricted.

Fig.~\ref{fig:main_ao} shows representative brain and knee slices from the DHS HyperRecon-L model. 
The two corresponding reconstructions are selected as follows. First, we densely sample uniformly from $[0,1]\times [0,1]$ to generate $100 \times 100$ reconstructions. Then, we filter out the reconstructions which are below some threshold PSNR value (we choose the 90th percentile). Finally, we choose the two reconstructions from this filtered set which are maximally separated by $\ell_2$ distance.

We notice that the two reconstructions are significantly dissimilar despite the similarity in PSNR value.
In the baseline setting, only one of these reconstructions would be available and finding another reconstruction would require training from scratch. 
With our model, users can search over the entire space of possible reconstructions and choose the one(s) they prefer.
This highlights the value of this model as a tool for the interactive selection of many diverse reconstructions for further use based on visual inspection.

\section{Conclusion and Future Work}
We presented a method for controllable image reconstruction using hypernetworks, which enables the interactive selection from a dense set of reconstructions at test-time. 
We highlight and address several issues related to hypernetwork training associated with the reconstruction setting, and demonstrate empirically that our method works on two datasets and on a variety of multi-term loss functions and reconstruction tasks.

We believe this work opens many interesting directions for further research. 
A straight-forward and promising future direction is to extend this to a higher number of loss coefficient hyperparameters. The challenges with this extension is two-fold. First, hypernetworks need to be designed to improve their expressivity for high-dimensional hyperparameter spaces. Second, since the space of possible reconstructions grows with additional hyperparameters, searching through all reconstructions becomes quickly intractable. Improved techniques for extracting the ``best" reconstructions to show to the user would be necessary.

Improving the expressivity and efficiency of hypernetworks is still under-explored. From Table 1, it can be seen that even the largest hypernetwork we experimented with requires three times less the number of parameters as compared to the combined parameters of all baseline models, with only a slight increase in training time. However, there may be more efficient yet expressive parameterizations of hypernetworks (for example, hypernetworks which only generate weights for certain layers of the main network) that can enable further compression.

The matched scaling of loss functions discussed in section~\ref{sec:matched} is an important open question. Since the optimal hyperparameter landscape is in general dependent on the scale of the loss functions and furthermore can affect the training dynamics, an automated and data-driven approach to loss scaling would be preferable. Insights may be gleaned from similar work in multi-task learning, where matching the contributions of losses associated with different tasks is important to the end performance~\citep{kendall2018multitask}. Other ideas associated with matching the scale of the gradients for each loss (instead of the scale of the loss itself) may also be useful.
\label{sec:conc}

\acks{This work was supported by NIH grants R01LM012719,
R01AG053949, the NSF NeuroNex grant 1707312, and the NSF
CAREER 1748377 grant (MS).}

%
\ethics{We used publicly available data that was anonymized/de-identified and originally collected with appropriate consents and approvals from the institutional IRBs and/or ethical review boards.}

\coi{We declare we don't have conflicts of interest.}

\bibliography{sample}

\newpage
\appendix 
\section*{Appendix}
\begin{figure*}[h]
\begin{centering}
\includegraphics[width=\textwidth]{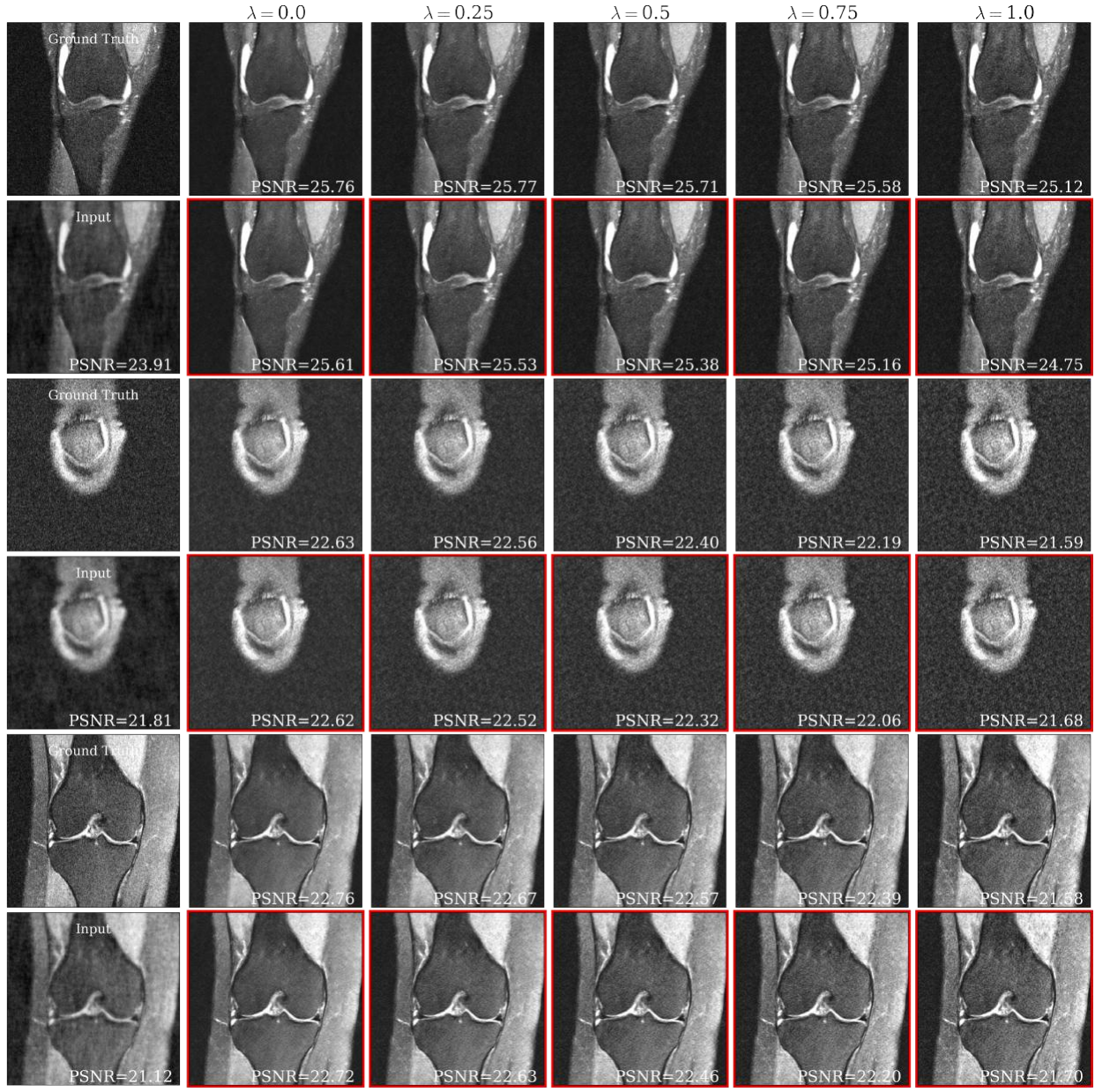}
\caption{Representative slices for knee on CS-MRI task with 8-fold under-sampling. Ground truth and input images are denoted in the first column. Top row are baseline Unet reconstructions with varying $\lambda$. Bottom row (red) are HyperUnet-L reconstructions with varying $\lambda$. PSNR values in reconstructions provide a basis of comparison between Unet and HyperUnet-L reconstructions of the same $\lambda$.}
\label{fig:csmri-slices}
\end{centering}
\end{figure*}

\begin{figure*}[t]
\begin{centering}
\includegraphics[width=\textwidth]{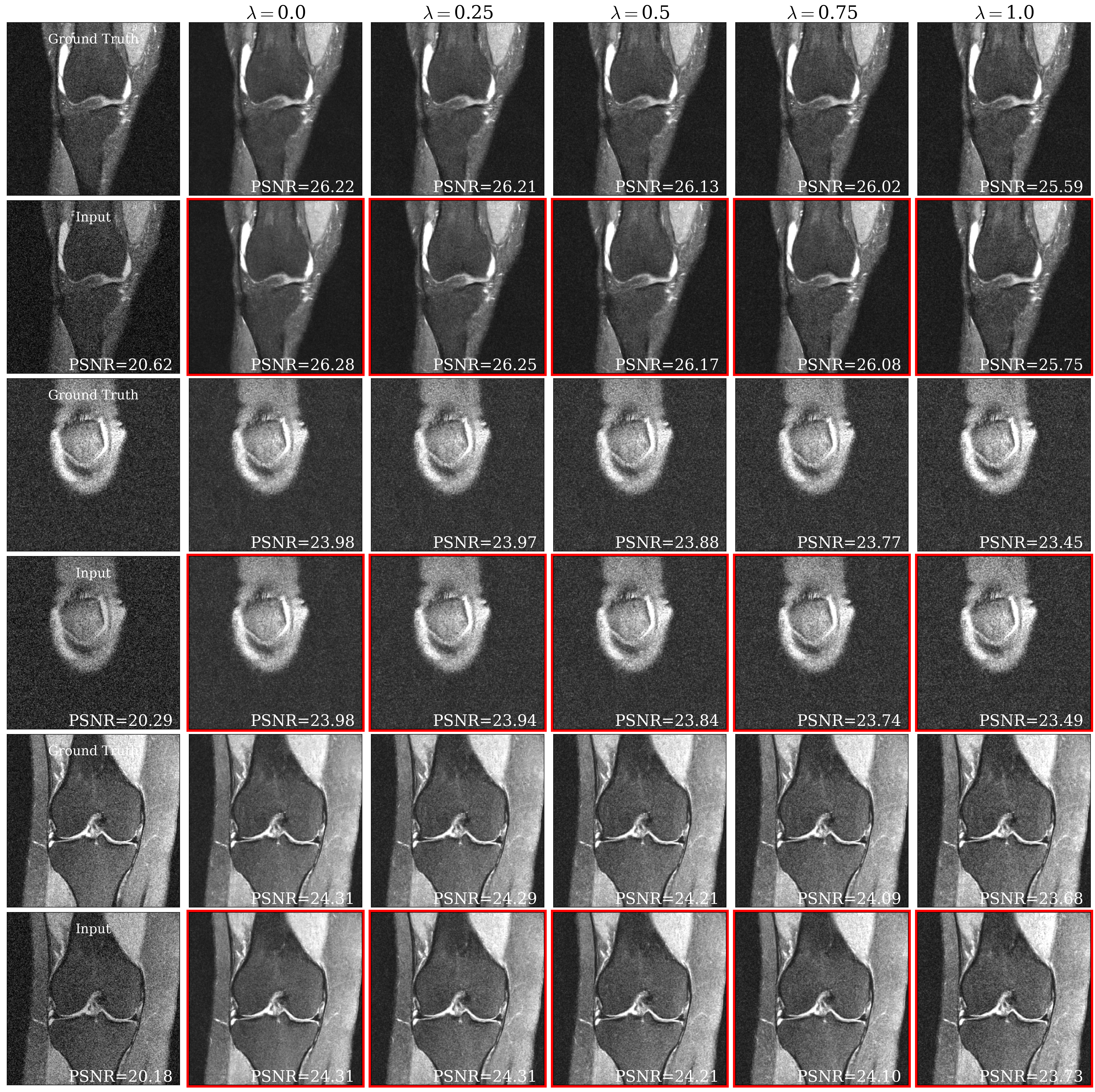}
\caption{Representative slices for knee on AWGN denoising task with noise standard deviation $\sigma=0.1$. Ground truth and input images are denoted in the first column. Top row are baseline Unet reconstructions with varying $\lambda$. Bottom row (red) are HyperUnet reconstructions with varying $\lambda$. PSNR values in reconstructions provide a basis of comparison between Unet and HyperUnet reconstructions of the same $\lambda$.}
\label{fig:denoise-slices}
\end{centering}
\end{figure*}
\begin{figure*}[t]
\begin{centering}
\includegraphics[width=\textwidth]{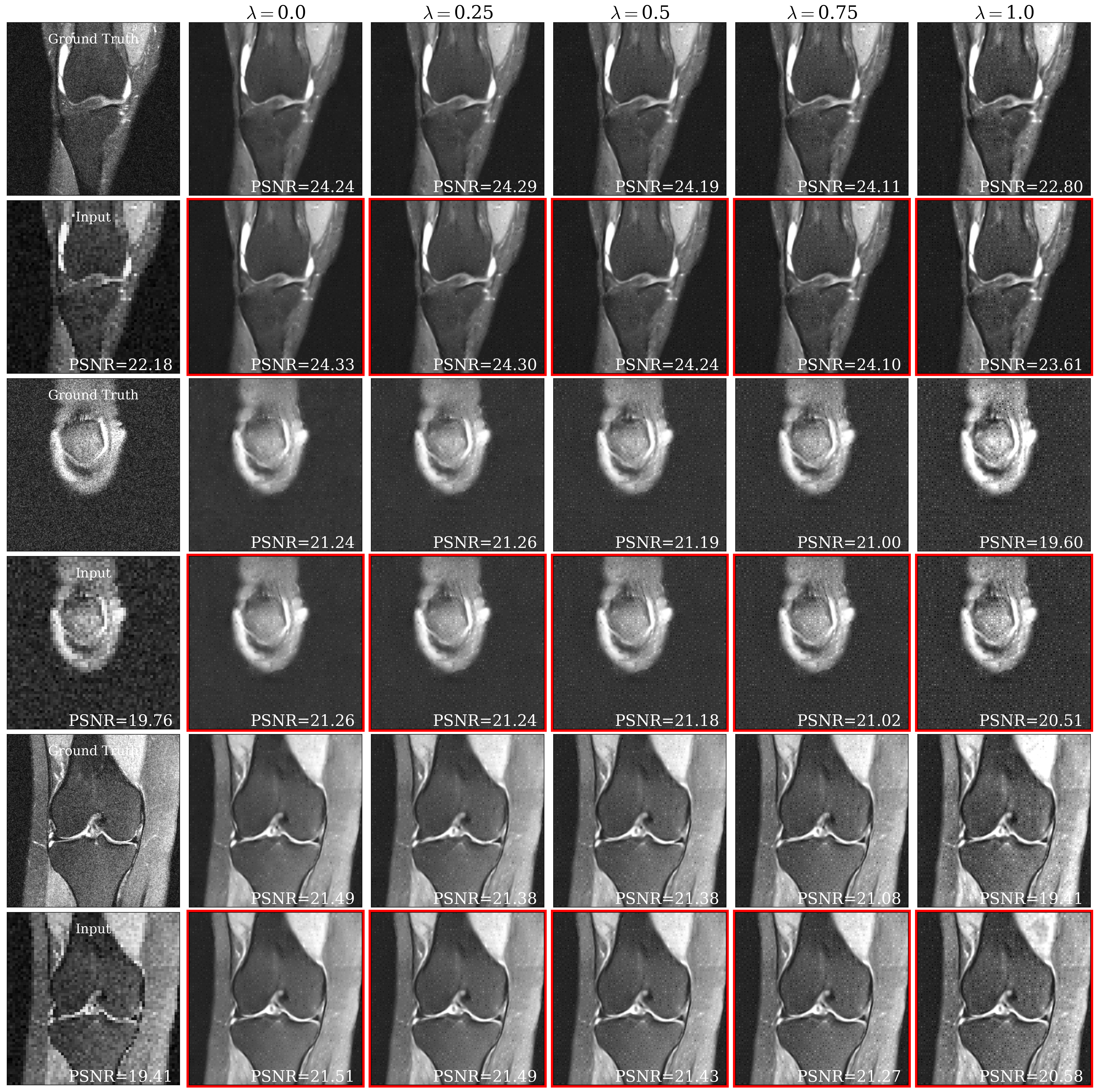}
\caption{Representative slices for knee on superresolution task at 4x downsampling. Ground truth and input images are denoted in the first column. Top row are baseline Unet reconstructions with varying $\lambda$. Bottom row (red) are HyperUnet-L reconstructions with varying $\lambda$. PSNR values in reconstructions provide a basis of comparison between Unet and HyperUnet-L reconstructions of the same $\lambda$.}
\label{fig:superres-slices}
\end{centering}
\end{figure*}

\begin{figure*}
\begin{centering}
\includegraphics[width=\textwidth]{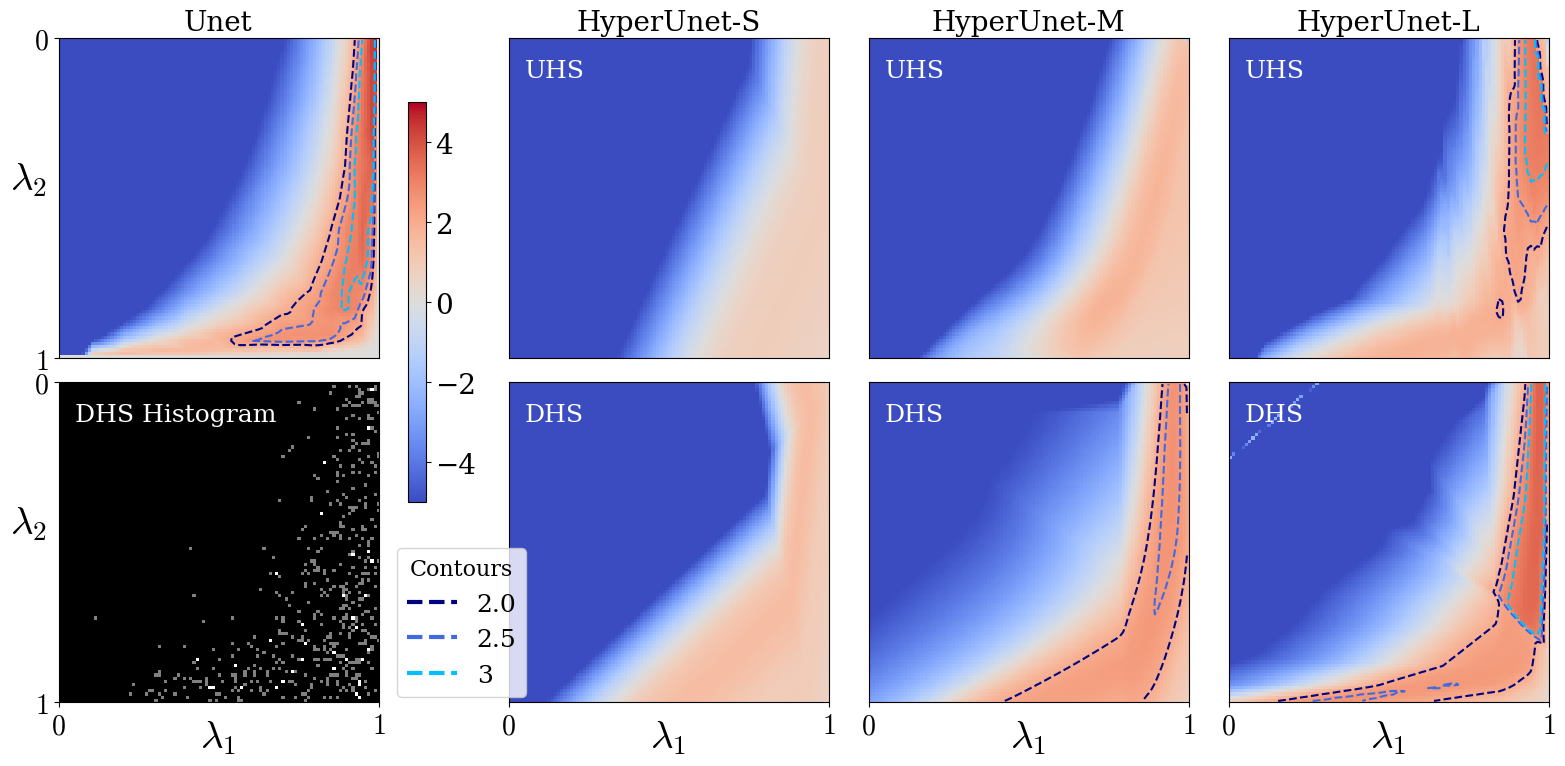}
\caption{rPSNR values for brain dataset over the hyperparameter support $[0,1]\times [0,1]$ for different hypernetwork capacity and sampling methods. 
The $x$-axis and $y$-axis denote the value of the hyperparameters $\lambda_1$ and $\lambda_2$, respectively. Contours denote level sets of fixed value (see legend). Stars in the landscapes denote maximum value. (Left) The top image depicts the baseline landscape. The bottom image shows an example histogram of hyperparameter values used for gradient computation during one epoch of training with the DHS strategy. (Right) The top and bottom row show the UHS and DHS model landscapes, respectively, for three different hypernetwork capacities.}
\label{fig:mask-compare-4x}
\end{centering}
\end{figure*}

\end{document}